\documentclass[10pt,twocolumn,letterpaper]{article}

\usepackage{iccv}              
\usepackage{float}

\usepackage{algorithm}
\usepackage{algorithmic}

\definecolor{tabfirst}{rgb}{1, 0.7, 0.7} 
\definecolor{tabsecond}{rgb}{1, 0.85, 0.7} 
\definecolor{tabthird}{rgb}{1, 1, 0.7} 

\usepackage{amssymb}
\usepackage{multirow}
\usepackage{colortbl}  

\usepackage{tablefootnote}
\usepackage{float}
\usepackage{booktabs}
\usepackage{pifont}

\definecolor{iccvblue}{rgb}{0.21,0.49,0.74}
\usepackage[pagebackref,breaklinks,colorlinks,allcolors=iccvblue]{hyperref}

\def\paperID{5008} 
\def\confName{ICCV}
\def\confYear{2025}

\title{GSV3D: Gaussian Splatting-based Geometric Distillation with Stable Video Diffusion for Single-Image 3D Object Generation}

\author{
Ye Tao$^{1}$, 
Jiawei Zhang$^{2}$, 
Yahao Shi$^{1}$, 
Dongqing Zou$^{2,3}$, 
and Bin Zhou$^{1,3}$ \\
$^1$State Key Laboratory of Virtual Reality Technology and Systems, Beihang University, \\$^2$SenseTime Research, $^3$PBVR
}

\let\oldtwocolumn\twocolumn
\renewcommand\twocolumn[1][]{%
    \oldtwocolumn[{#1}{
    \begin{center}
           \includegraphics[width=1\linewidth]{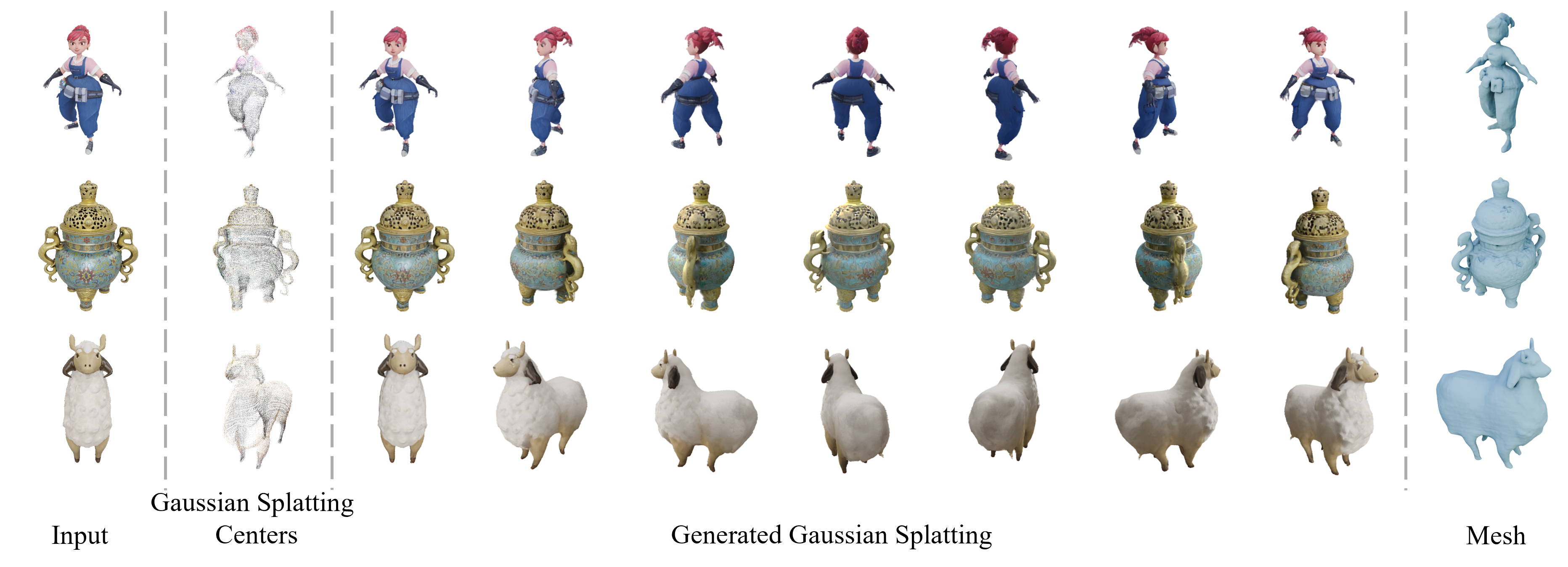}
           \vspace{-24pt}
           \captionof{figure}{
          GSV3D utilizes Stable Video Diffusion and Gaussian Splatting decoder to generate 3D model from a single image.
          Meanwhile, the Gaussian Splatting-based geometric distillation is used to enhance the 3D consistency.}
           \label{fig:teaser}
        \end{center}
    }]
}

\begin{document}

\maketitle

\begin{abstract}

Image-based 3D generation has vast applications in robotics and gaming, where high-quality, diverse outputs and consistent 3D representations are crucial.
    However, existing methods have limitations: 3D diffusion models are limited by dataset scarcity and the absence of strong pre-trained priors, while 2D diffusion-based approaches struggle with geometric consistency.
    We propose a method that leverages 2D diffusion models' implicit 3D reasoning ability while ensuring 3D consistency via Gaussian-splatting-based geometric distillation.
    Specifically, the proposed Gaussian Splatting Decoder enforces 3D consistency by transforming SV3D latent outputs into an explicit 3D representation.
    Unlike SV3D, which only relies on implicit 2D representations for video generation, Gaussian Splatting explicitly encodes spatial and appearance attributes, enabling multi-view consistency through geometric constraints.
    These constraints correct view inconsistencies, ensuring robust geometric consistency.
    As a result, our approach simultaneously generates high-quality, multi-view-consistent images and accurate 3D models, providing a scalable solution for single-image-based 3D generation and bridging the gap between 2D Diffusion diversity and 3D structural coherence.
    Experimental results demonstrate state-of-the-art multi-view consistency and strong generalization across diverse datasets.
    The code will be made publicly available upon acceptance.
\end{abstract}    
\section{Introduction}

Image-based 3D generation has emerged as a crucial area of research with transformative applications in robotics, gaming, virtual reality, and other areas where accurate 3D modeling from limited information can bring substantial benefits. 
However, achieving this goal remains challenging due to the intrinsic limitation of 2D diffusion models in ensuring geometric consistency and the reliance on 3D diffusion models on limited datasets, leading to fidelity issues and monotonous textures.
Existing purely 3D diffusion models are often limited by the scarcity of datasets, which limits their ability to produce diverse and realistic results~\cite{zeng2022lion,nichol2022point,gao2022get3d,Liu2023MeshDiffusion,jun2023shap,li2023diffusion,zhang2024clay, li2024craftsman, lan2024ga}.
This limitation restricts the generalization ability of these models and makes it difficult for them to capture complex details across various object types. 
Conversely, 2D diffusion-based methods have shown substantial diversity in generated results due to their reliance on rich 2D datasets~\cite{rombach2022high}.
However, maintaining geometric consistency across multiple views remains a significant challenge. 
Although 2D diffusion models often struggle to maintain geometric consistency, they possess the implicit ability to generate consistent structures. 
Some recent approaches have attempted to address this by directly fine-tuning 2D diffusion models with additional attention mechanisms to encourage multiview consistency~\cite{shi2023zero123++,shi2023mvdream,wang2023imagedream,yang2024hi3d,voleti2024sv3d}. 
Yet, these methods primarily apply implicit constraints that lack the robustness needed for consistent geometric alignment, often resulting in breakdowns in multiview consistency. 
The absence of explicit geometric constraints frequently leads to discrepancies across views, ultimately compromising the accuracy and usability of the generated 3D models.

To address these issues, we propose a framework that combines the diversity of 2D diffusion with the geometric consistency required for 3D generation.
Our approach builds on SV3D~\cite{blattmann2023stable}, an advanced diffusion model optimized for generating coherent video sequences that rotate around an object from single images. 
To further enhance multi-view consistency, we introduce geometric distillation, a strategy that transforms multi-view latents into a structurally aligned 3D representation to enforce cross-viewpoint coherence.
Unlike standard latent diffusion processes, which operate purely in 2D space, our method applies a Gaussian Splatting-based transformation to distill multi-view information into a unified 3D-aware representation.
By representing latents in 3D, this approach amplifies geometric inconsistencies between different latents, providing an opportunity to incorporate additional supervision, such as depth information, to improve their consistency.
This ensures more accurate and coherent results across views. 
Consequently, our approach not only overcomes the geometric inconsistencies of 2D diffusion models but also directly generates 3D results without requiring additional reconstruction steps. 
Furthermore, compared to 3D diffusion models, our method achieves higher fidelity and more realistic, diverse textures, providing a robust and reliable solution for high-quality, geometrically consistent multiview generation.

In summary, our contributions are as follows:

\begin{itemize}

\item We propose a latent decoder trained to extract 3D Gaussian Splatting representations directly from multi-view latents generated by the diffusion model, enabling direct 3D model generation. 

\item We introduce a method that leverages Gaussian Splatting for geometric distillation within the SVD framework, enforcing structural coherence to enhance multi-view consistency and improve 3D reconstruction fidelity from a single image.

\item We validate our approach through extensive experiments, showcasing its superior performance and generalization capabilities in generating consistent 3D objects from 2D inputs.
\end{itemize}

\section{Related Works}
\begin{figure*}[ht]
  \centering
   \includegraphics[width=1\linewidth]{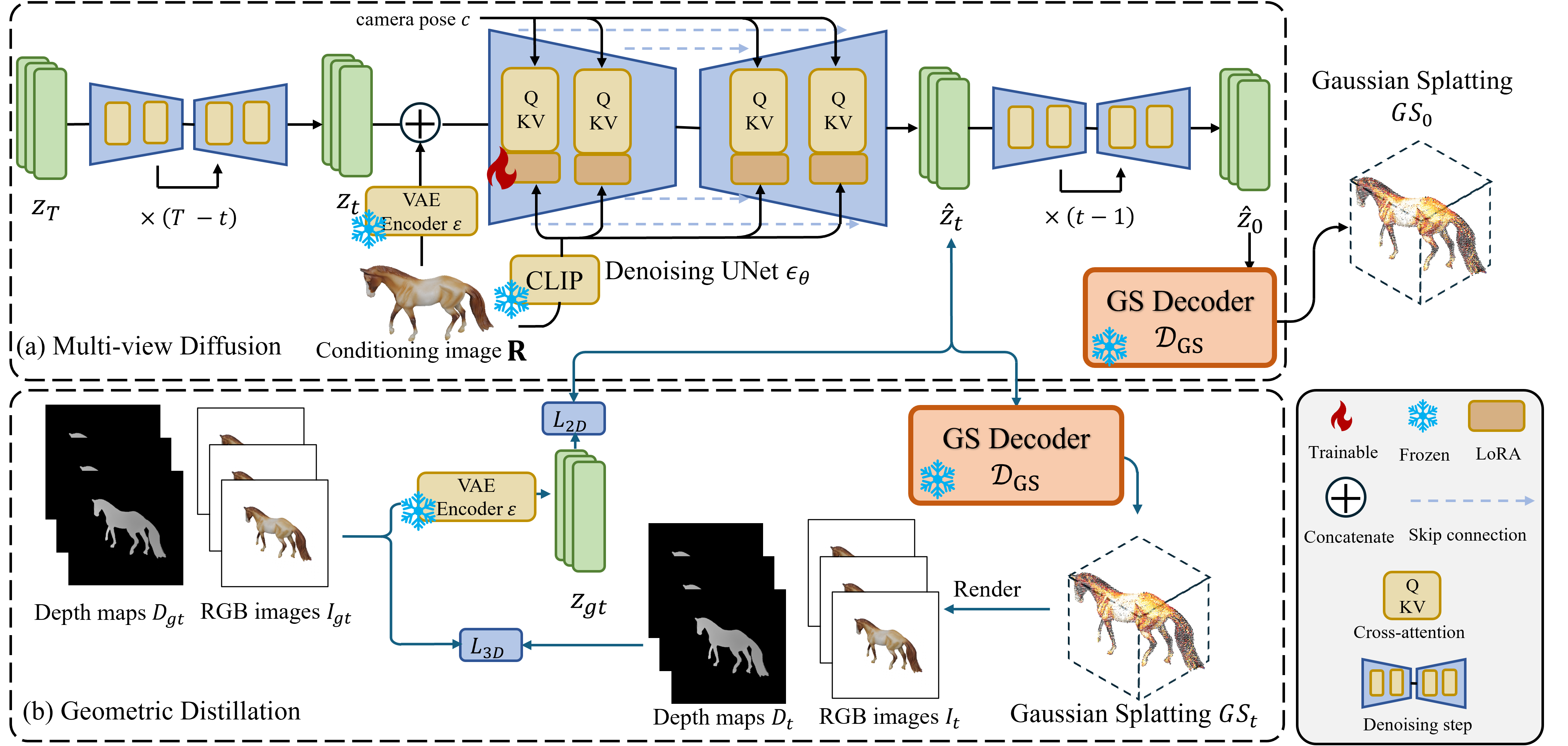}
   \vspace{-20pt}
   \caption{\textbf{Overview of GSV3D Training and Inference Pipeline.} 
    During inference, given an initialized noise latent \(z_T\), an input image \(\mathbf{R}\) and its corresponding camera pose \(c\), the approach follows a structured pipeline: 
    At each step of the sampling, the input image is encoded using VAE \(\mathcal{E}\), and its latent representation is concatenated with input noisy multi-view latents \(z_t\). 
    Simultaneously, the image \(\mathbf{R}\) is encoded via CLIP and, along with the camera pose \(c\), integrated into the diffusion process via cross-attention. 
    For the last step, the multi-view denoised latents \(\hat{z_0}\) are processed by the Gaussian Splatting Decoder to reconstruct a 3D representation \(GS_0\), as shown in (a).
    During training, the geometric distillation process utilizes the Gaussian Splatting Decoder with 3D constrain to distill geometric knowledge from the multi-view latent representations \(\hat{z_t}\) generated by the multi-view diffusion model at each diffusion step, which then serves as supervision \(\mathcal{L}_{\text{3D}}\) to guide the learning of 3D geometry.
    At the same time, the multi-view latent loss \(\mathcal{L}_{\text{2D}}\) is imposed in latent space, as shown in (b).}
    \vspace{-8pt}
   \label{fig:pipeline}
\end{figure*}
\subsection{Native 3D Generative Models}

Recent advancements in 3D generative modeling have seen a rise in native 3D generation methods, which directly create 3D representations. 
These approaches are diverse, as 3D data can be represented in various forms, Signed Distance Fields (SDF), meshes, point clouds, NeRF, and Gaussian Splatting. 
However, it's important to highlight that certain 3D representations, such as point clouds~\cite{zeng2022lion, nichol2022point},  SDF~\cite{li2023diffusion, zhang2024clay, li2024craftsman}, and meshes~\cite{Liu2023MeshDiffusion, chen2024meshanything}, are limited to capturing geometry and do not support texture generation. 
As a result, additional work is required to generate textures separately for these models. 
On the other hand, representations like NeRF~\cite{hong2023lrm, jun2023shap} and Gaussian Splatting~\cite{roessle2024l3dg, lan2024ga, zou2024triplane} are more integrated, as they can model both geometry and texture simultaneously, which makes them more powerful in generating visually compelling 3D content. 
Despite these advancements, the scarcity and monotony of available 3D datasets~\cite{deitke2023objaverse, deitke2024objaverse} remain significant challenges. 
This limitation means that even with advanced methods, generated geometries and textures often lack high fidelity and realism, reinforcing the need for further research to address these issues. Overall, while these native 3D generation methods show great promise, they still face challenges related to dataset diversity, fidelity, and the balance between geometry and texture.

\subsection{2D Generative Models for 3D Generation}
While native 3D generation methods still face challenges due to the scarcity of diverse 3D datasets, 2D diffusion models, benefiting from abundant 2D data~\cite{schuhmann2022laion}, have shown exceptional diversity and realism in generated results~\cite{rombach2022high}. 
However, a major issue with 2D-generated content is the geometric inconsistency between multiple images, making it difficult to reconstruct a coherent 3D model from them. 
To address this, early works like DreamFusion~\cite{poole2022dreamfusion} proposed using a Score-Distilled (SDS) loss to optimize a randomly initialized NeRF field, essentially distilling 3D structure from 2D diffusion models.
While this approach demonstrated promising results in 3D generation, it was highly time-consuming, with some methods~\cite{wang2024prolificdreamer} taking hours to generate a single object. In response to the inefficiency of SDS-based methods, later research focused on equipping 2D diffusion models with some degree of 3D awareness. 
For example, Zero-1-to-3~\cite{liu2023zero} integrated camera pose information into the conditioning process of the 2D model, allowing it to generate images from novel viewpoints. 
Methods~\cite{shi2023mvdream, voleti2024sv3d, wang2023imagedream, shi2023zero123++, yang2024hi3d} leverage attention mechanisms to enhance multi-view consistency, either by enforcing geometric alignment across multiple views or by transferring temporal consistency from videos to 3D object generation, to ensure that the generated images adhere to the same 3D structure. Despite these advancements, the attention mechanisms that enhance multi-view consistency primarily achieve geometric alignment in appearance rather than understanding true 3D structures. 
These models may generate visually consistent views but still struggle with robust 3D reconstruction. 
When applied to tasks such as object reconstruction from multiple views, these methods often reveal geometric inconsistencies, highlighting a significant gap in the ability of current 2D diffusion models to fully comprehend and generate 3D data.


\section{Method}
\begin{figure*}[ht]
  \centering
   \includegraphics[width=1\linewidth]{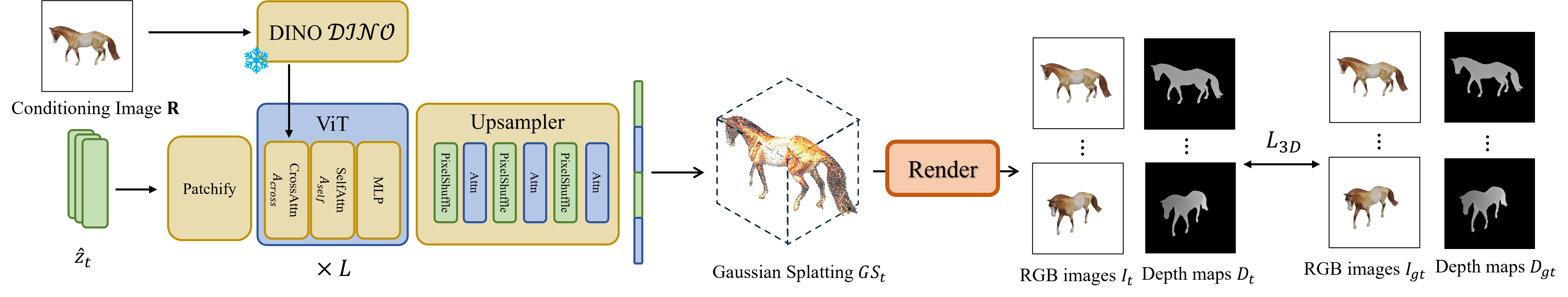}
   \vspace{-24pt}
   \caption{\textbf{Overview of the Gaussian Splatting Decoder pipeline. }
   During both geometric distillation and GSV3D inference, the conditioning image \(\mathbf{R}\) is first processed by a pre-trained DINO encoder \(\mathcal{DINO}\) to extract features, which are then integrated into the ViT via cross-attention.
   The ViT processes the latents \(\hat{z_{t}}\) generated by GSV3D. 
    The output of the ViT is passed through an upsampler before being converted into a 3D Gaussian Splatting representation. 
   In order to train the Gaussian Splatting Decoder, the latents \(\hat{z_{t}}\) are substituted with the latents \(z_{input}\) encoded from the input multi-view images \(I_{input}\) using a pre-trained VAE encoder \(\mathcal{E}\).
   These input multi-view images consist of \(N\) images, each corresponding to a viewpoint uniformly distributed around the object.
   }
   \vspace{-8pt}
   \label{fig:gs-decoder}
\end{figure*}
\subsection{Overview}
\label{sec:overview}
To leverage the powerful priors of 2D diffusion models, generating a 3D representation \(GS_0\) from a single image \(\mathbf{R}\) requires synthesizing realistic consistent multi-view observations. 
SV3D~\cite{voleti2024sv3d} is a multi-view diffusion model that can generate 360-degree multi-view images around an object by modeling temporal dependencies (Section~\ref{sec:svd}).
While it produces visually realistic results, the absence of explicit 3D constraints results in geometrically inconsistent reconstructions.
To address this, we introduce a Gaussian Splatting Decoder, which transforms the multi-view latents into an explicit 3D structure (Section~\ref{sec:gs-decoder}). 
These multi-view latents are intermediate representations that capture and encode the features of multi-view images generated by the multi-view diffusion model.
This decoder provides a learnable 3D representation, better capturing the underlying geometry.
With the above Gaussian Splatting Decoder, we propose a framework GSV3D, which incorporates the Gaussian Splatting Decoder into its training pipeline to enhance the 3D awareness of the multi-view diffusion model.
This integration effectively enables the multi-view diffusion model to generate geometrically consistent multi-view latents by incorporating 3D constraints through geometric distillation (Section~\ref{sec:train}).
Finally, we use the Gaussian Splatting Decoder to directly convert the multi-view latents generated by GSV3D to Gaussian Splatting, resulting in a 3D representation. An overview is shown in Figure~\ref{fig:pipeline}.

\subsection{Multi-view Diffusion Model for 3D Generation}
\label{sec:svd}
SV3D is a multi-view diffusion model proposed in prior work which is built upon the Stable Video Diffusion~\cite{blattmann2023stable} (SVD) framework.
It is capable of generating a 360-degree rotation video around an object from a single image.
This capability is achieved by fine-tuning the SVD model on a 3D dataset and incorporating a camera pose control mechanism that controls the viewpoint of the generated images.
In this section, we provide a brief overview of the SV3D architecture as described in the original work, which serves as a foundation of our approach.

The conditioning image \(\mathbf{R}\) is encoded into two distinct representations using pre-trained encoders: a VAE encoder and a CLIP encoder.
The VAE encoder \(\mathcal{E}\) captures geometric and appearance features, producing a latent representation that is concatenated with the noisy multi-view latents \(z_t\) at each denoising step.
On the other hand, the CLIP encoder extracts semantic information, which is integrated into the denoising process through a cross-attention mechanism.
This dual-encoding strategy ensures that both low-level geometric details and high-level semantic context are effectively utilized during generation.
Additionally, to control the generated view, the camera pose \(c\) for each generated image is encoded using cosine encoding and incorporated into the cross-attention mechanism.
The UNet \(\epsilon_\theta\) is a denoising model, which utilizes cross-frame attention to ensure spatial consistency in the denoised multi-view latents.

During inference, the process starts from pure noise \(z_T\) in the latent space, and the denoising model \(\epsilon_{\theta}\) iteratively refines it through a series of denoising steps.
The final multi-view latent representations \(\hat{z_0}\) are then decoded by the VAE decoder to reconstruct the output images, which consist of \(N\) images corresponding to distinct viewpoints that are approximately uniformly distributed around the object.
However, although the generated images appear visually plausible, their lack of geometric consistency poses significant challenges for accurate 3D reconstruction, which heavily relies on precise geometric consistency.

\subsection{Gaussian Splatting Decoder}
\label{sec:gs-decoder}

To achieve consistent multi-view results in GSV3D, we propose the Gaussian Splatting Decoder.
This decoder extracts the implicit geometric information from the multi-view latents generated by the multi-view diffusion model.
Through the geometric distillation process(detailed in Section~\ref{sec:train}), it introduces 3D constraints into GSV3D, enhancing its ability to produce geometrically consistent multi-view latents.
Additionally, the Gaussian Splatting Decoder can convert the multi-view latents generated by GSV3D into a Gaussian Splatting representation directly to obtain the final 3D result.

To achieve this, we adopt a structure inspired by GRM~\cite{xu2024grm}, where a VisionTransformer (ViT)~\cite{alexey2020image} is used to extract features from latents, followed by an Upsampler to generate a high-resolution Gaussian Splatting output, as shown in Figure~\ref{fig:gs-decoder}.

Specifically, latents \(\hat{z_{t}}\) are processed by a ViT with $L=16$ layers, which extracts hierarchical features through self-attention mechanisms \(A_{self}\).
These mechanisms capture both local and global dependencies by operating on patches of the latent representations.
To improve the quality of generated Gaussian Splatting, we also introduce a cross-modal feature fusion strategy. 
This strategy integrates high-level semantic features from conditioning image \(\mathbf{R}\) to enhance the latent space representation.
Specifically, we extract global structural cues using a pre-trained DINO~\cite{caron2021emerging, oquab2023dinov2} model \(\mathcal{DINO}\), which provides a feature representation denoted as \(\mathbf{F}_{\text{DINO}}\):
\begin{equation}
    \mathbf{F}_{\text{DINO}} = \mathcal{DINO}(\mathbf{R}).
\end{equation}
The extracted DINO features \(\mathbf{F}_{\text{DINO}}\) capture important semantic context, which helps guide the model in generating more coherent and structurally consistent reconstructions.
To incorporate these global cues into ViT processing, we introduce a cross-attention mechanism that allows the DINO features \(\mathbf{F}_{\text{DINO}}\) to attend to the intermediate features of ViT \(\mathbf{F}_{\text{ViT}}\).
This mechanism is integrated into each ViT layer and operates as:
\begin{equation}
    \mathbf{F}_{\text{ViT}} \leftarrow \mathcal{A}_{\text{cross}}(\mathbf{F}_{\text{ViT}}, \mathbf{F}_{\text{DINO}}),
\end{equation}
where \( \mathcal{A}_{\text{cross}}(\cdot, \cdot) \) denotes the cross-attention operation.
This integration allows the latent space to be enriched with global structural priors, which improves the quality of the Gaussian Splatting reconstruction.
Finally, these extracted features are processed by the Upsampler to generate the final Gaussian Splatting reconstruction \(GS\).
The Upsampler consists of a series of pixel shuffle layers and transformer layers that gradually increase the spatial resolution of the features while reducing the number of channels.
The output of the Upsampler is a high-resolution 3D Gaussian Splatting representation.

\subsection{Training Strategy}
\label{sec:train}

\subsubsection{Training Gaussian Splatting Decoder}
\label{sec:train-decoder}

In order to train the Gaussian Splatting Decoder, we render \(Q\) images for each of the \(P\) objects in the training set, with these images evenly distributed around the object.
For each object, \(N\) images \(\{I_{input}\}_{n=1}^N\) are uniformly selected from \(Q\) images as input views.
All \(Q\) images serve as ground-truth images \(\{I_{gt}\}_{q=1}^Q\) to supervise the corresponding object.
Unlike in the geometric distillation process, we need to generate multi-view latents \(z_{input}\) to replace the multi-view latents \(\hat{z_t}\) as the input of the Gaussian Splatting Decoder.
These latents \(z_{input}\) are obtained by encoding the input multi-view images \(I_{input}\) using a pre-trained VAE \(\mathcal{E}\).
Meanwhile, the first input image \(I_{input}^0\) is used as the conditioning image \(\mathbf{R}\).

To optimize the Gaussian Splatting Decoder, we render the RGB images \(I_{\text{output}}\) and depth maps \(D_{\text{output}}\) from the Gaussian Splatting representation \(GS_{\text{output}}\) generated by the model, as shown in Figure~\ref{fig:gs-decoder}.
Specifically, we use \(I_{\text{gt}}^{(p,q)}\), \(I_{\text{output}}^{(p,q)}\), \(D_{\text{gt}}^{(p,q)}\), and \(D_{\text{output}}^{(p,q)}\) to refer to the \(q\)-th ground-truth and rendered image/depth map for the \(p\)-th object.
Notably, we render both the input views and the novel views to ensure a comprehensive optimization of the performance of the decoder across different perspectives.
These rendered outputs are then used in the loss calculation, which involves a combination of loss functions.
The RGB loss \( \mathcal{L}_{\text{rgb}} \) is computed as the pixel-wise mean squared error (MSE) between the rendered RGB images \( I_{\text{t}} \) and the corresponding ground truth images \( I_{\text{gt}} \):
\begin{equation}
    \mathcal{L}_{\text{rgb}} = \frac{1}{P \cdot Q} \sum_{p=1}^{P} \sum_{q=1}^{Q} \left\| I_{\text{output}}^{(p,q)} - I_{\text{gt}}^{(p,q)} \right\|_2^2.
\end{equation}
To further constrain the geometry of the generated structure, we apply depth loss \(\mathcal{L}_{\text{depth}}\), which compares the rendered depth maps \(D_{\text{output}}\) with the ground-truth depth maps \(D_{\text{gt}}\):
\begin{equation} 
\mathcal{L}_{\text{depth}} = \frac{1}{P\cdot Q} \sum_{p=1}^{P}\sum_{q=1}^{Q} || D_{\text{output}}^{(p,q)} - D_{\text{gt}}^{(p,q)} ||_2^2.
\end{equation}
The total loss function \( \mathcal{L}_{\text{3D}} \) is defined as a weighted sum of multiple components, each contributing to different aspects of the reconstruction quality:
\begin{equation}
\label{eq_3d}
    \mathcal{L_{\text{3D}}} = \mathcal{L}_{\text{rgb}} + \lambda_{\text{depth}}\mathcal{L}_{\text{depth}},
\end{equation}
The weight parameter\(\lambda_{\text{depth}}\) is introduced to balance the contributions of perceptual quality and depth supervision.

\subsubsection{Geometric Distillation Process for GSV3D}
\label{sec:svd-gs}

In order to train GSV3D, we utilize the same method to generate the training dataset as we mentioned in Section~\ref{sec:train-decoder}.
We choose the first ground-truth frame \(I_{input}^0\) as the conditioning image \(\mathbf{R}\) for GSV3D.
The ground-truth latent representations \(z_{gt}\) are encoded from the ground-truth multi-view images \(I_{gt}\) using the pre-trained VAE encoder \(\mathcal{E}\).

While the multi-view diffusion model inherently possesses potential 3D reasoning capabilities, as it is able to generate multi-view perspectives that rotate around the object from a single 2D input image, its full 3D potential can only be unlocked with a framework that directs the model to better understand and leverage 3D structural information.
To enhance the 3D awareness of the multi-view diffusion model, we introduce a geometric distillation framework that implicitly encodes 3D structural constraints into the generative process.
It ensures that the model learns to reason about spatial relationships and multi-view consistency in a way that respects geometric constraints.
The core of this approach is the Gaussian Splatting Decoder, which acts as the 3D-Aware Branch.
At each time step \(t\), Gaussian noise is added to \(z_{gt}\), producing noisy latents \(z_t\) according to a predefined noise schedule.
These noisy latents \(z_t\) are then processed by the multi-view diffusion model, which outputs denoised latents \(\hat{z_t}\).
The denoised latents, along with the conditioning image \(\mathbf{R}\), are then passed through the Gaussian Splatting Decoder.
We freeze the Gaussian Splatting Decoder and distill the multi-view diffusion model by attaching a LoRA~\cite{hu2021lora} (Low-Rank Adaptation) module, which allows us to update the parameters of the multi-view diffusion model while preserving the information necessary for the 3D constrain in the Gaussian Splatting Decoder, as shown in Figure~\ref{fig:pipeline}.

To distill the multi-view diffusion model, we apply the 3D loss \(\mathcal{L}_{\text{3D}}\) in Eq.~\ref{eq_3d} to update the diffusion model.
Additionally, we retain the original multi-view latents denoising loss, denoted as $\mathcal{L}_{\text{2D}}$. 
This loss is used to optimize the 2D diffusion model by comparing the generated noisy latents with the clean latent targets:
\begin{equation}
    \mathcal{L}_{\text{2D}} = \mathbb{E}_{\mathbf{R}, z_t, z_{gt}} \left[ \| \epsilon_\theta(z_t;\mathbf{R}, t) - z_{gt} \|_2^2 \right],
\end{equation}
where \(\mathbb{E}_{\mathbf{R}, z_t, z_{gt}}\) denotes the expectation over the joint distribution of training images \(\mathbf{R}\), their corresponding noisy latents \(z_t\) and ground-truth latnets \(z_{gt}\), \(\epsilon_\theta(z_t; \mathbf{R}, t)\) is the output of the denoising network.

The total loss function, combining both the multi-view latents denoising loss ($\mathcal{L}_{\text{2D}}$) and the 3D consistency loss ($\mathcal{L}_{\text{3D}}$), is given by:

\begin{equation}
\mathcal{L}_{\text{distill}} = \mathcal{L}_{\text{2D}} + \lambda_{\text{3D}}\mathcal{L}_{\text{3D}},
\end{equation}
where \( \lambda_{\text{3D}} \) is a hyperparameter that controls the relative weighting of the 2D and 3D losses.

\section{Experiments}
\begin{figure*}[ht]
  \centering
   \includegraphics[width=1\linewidth]{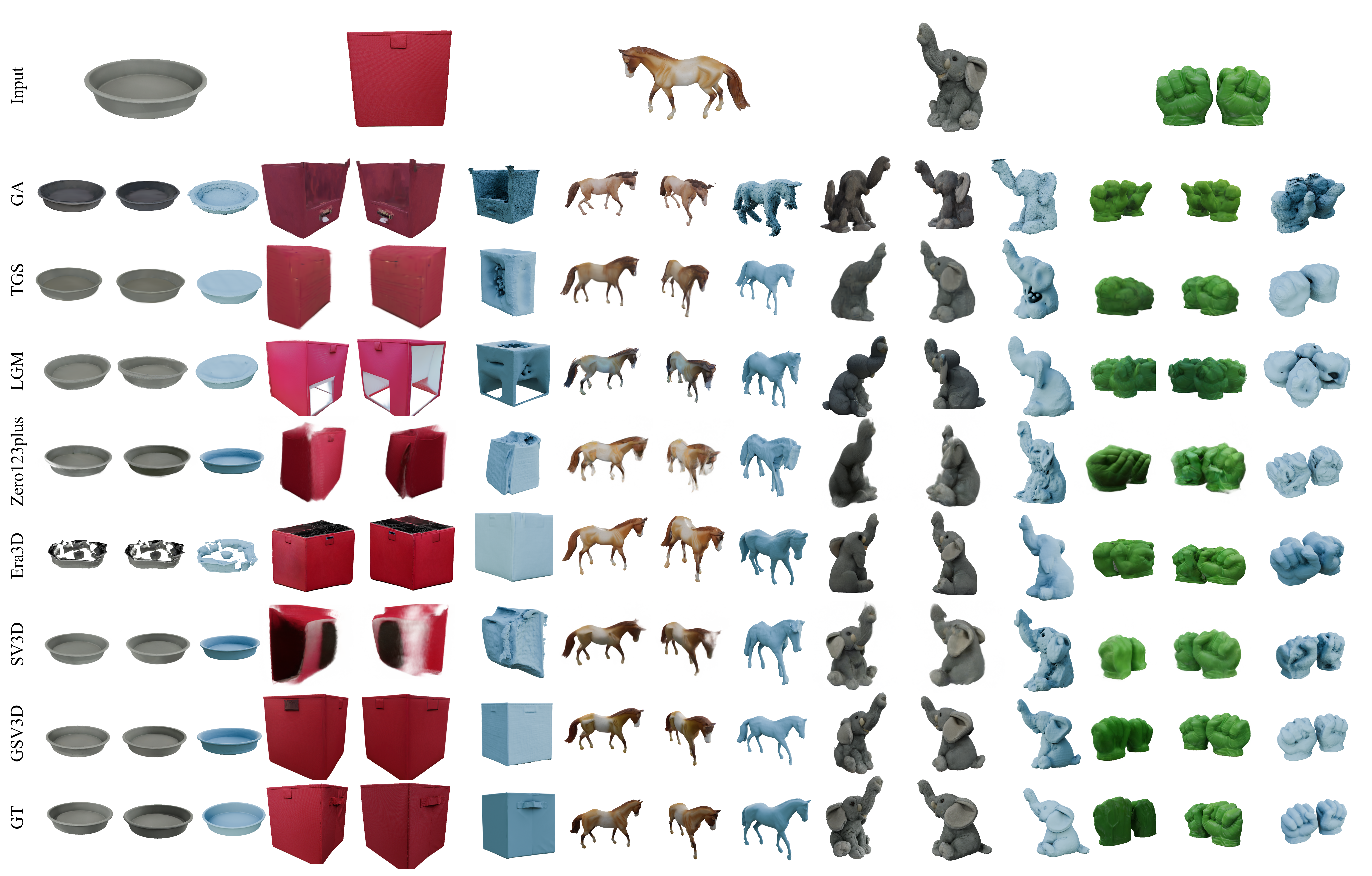}
   \vspace{-16pt}
   \caption{Performance comparison between our GSV3D and other state-of-art methods. 
   GA and TGS are abbreviations for GaussianAnything and TriplaneGaussian, respectively. 
   For each example, the first two columns display two different rendering views of the generated 3D representation, while the third column shows the rendered image of the extracted Mesh.}
   \vspace{-12pt}
   \label{fig:compare}
\end{figure*}

\subsection{Experimental Settings}
\noindent \textbf{Datasets.} For training, we collect \(P=1\times10^5\) high-quality 3D models from Objaverse~\cite{deitke2023objaverse} dataset.
For each model, we render \(Q=84\) sets of RGB images and depth maps.
These images capture views of the object from different perspectives, where the azimuth angles are uniformly distributed around the object.
While the elevation angle is fixed for each individual object, it is randomly sampled between  -5 and 30 degrees across different objects.
From these 84 images, we select \(N=16\) images at regular intervals to serve as the training input.
For evaluation, we select a standardized subset of 100 randomly chosen models from the Google Scanned Objects (GSO)~\cite{downs2022google} dataset.
For each model, we render 21 RGB images with a fixed elevation angle, which is randomly chosen between -5 and 30 degrees, and azimuth angles uniformly distributed across 360 degrees. 
The first rendered image of each object is designated as the input for GSV3d.
The rest of the images are used for evaluation.
Also, we sample 4,096 points from the mesh of each object to facilitate 3D evaluation.

\vspace{4pt}
\noindent\textbf{Metrics.} For evaluation, we assess the quality of appearance and geometry.
For appearance evaluation, we render images from 21 viewpoints of the generated 3D representations. 
We employ various metrics: PSNR, SSIM, LPIPS, KID~\cite{binkowski2018demystifying}, FID~\cite{heusel2017gans}, and CLIP-IQA~\cite{wang2022exploring} for appearance quality.
For the geometry evaluation, we extract meshes from the generated 3D representations and then sample points from these meshes to compute the Chamfer Distance (CD), Intersection over Union (IoU), and F-Score with the ground truth point cloud, which measures geometric accuracy.
We also conduct a user study to evaluate the perceptual realism and structural plausibility of the outputs.

\vspace{4pt}
\noindent\textbf{Training Details.} Our training process consists of two stages:
\begin{itemize}
    \item \textbf{Stage One:} We begin by training a Gaussian Splatting Decoder, as we mentioned in Section~\ref{sec:gs-decoder}.
    We train this model from scratch to obtain 3D representations from multi-view latents.
    This stage is trained for 80,000 steps with a batch size of 128.

    \item \textbf{Stage Two:}
    We integrate the Gaussian Splatting Decoder into the multi-view diffusion model in GSV3D to perform geometric distillation, as mentioned in Section~\ref{sec:svd-gs}.
    The Gaussian Splatting Decoder is initialized with the first-stage trained model, while the UNet, VAE encoder, and CLIP encoder are initialized with the pre-trained SV3D model.
    During distillation, the parameters of the Gaussian Splatting Decoder, the UNet, VAE encoder, and CLIP encoder are frozen, with only the LoRA parameters, which are incorporated into the UNet model, being updated.
    This stage is trained for 40,000 steps with a reduced batch size of 32.
\end{itemize}
In both stages, we use AdamW~\cite{loshchilov2017decoupled} with a learning rate of $1 \times 10^{-5}$ to optimize the model. The hyperparameter \(\lambda_{\text{3D}}\) and \(\lambda_{\text{depth}}\) is set to 1.5 and 0.2, respectively.
The experiments are conducted on 4 NVIDIA A100-80G GPUs.

\begin{table*}[t]
\centering
{
\label{tab:tensoir}
\resizebox{\linewidth}{!}{
\begin{tabular}{c|l|cccccc|ccc|cc}
\toprule
\multirow{2}{*}{Type}&\multirow{2}{*}{Method} & \multicolumn{6}{c|}{Appearance Quality} & \multicolumn{3}{c|}{Geometry Quality} & \multicolumn{2}{c}{User Study}          \\
                        &  &PSNR$\uparrow$   & SSIM$\uparrow$  & LPIPS$\downarrow$   & FID$\downarrow$ & KID$\downarrow$ & CLIP-IQA$\uparrow$ & CD$\downarrow$ & IoU $\uparrow$ & F-Score$\uparrow$ & App. Score$\uparrow$ & Geo. Score$\uparrow$
                        \\ 
\midrule
\multirow{2}{*}{{3D}}&GA\cite{lan2024ga}    & 15.201     & 0.834      & 0.039    & 95.47     & 1.17      & 0.805 & \cellcolor{tabthird}0.197    & \cellcolor{tabthird}0.502 & 0.303 & 3.154 & \cellcolor{tabsecond}5.000\\
        &TGS\cite{zou2024triplane}    & \cellcolor{tabsecond}18.874     & \cellcolor{tabsecond}0.872      & \cellcolor{tabsecond}0.032    & 85.25     & 1.28      & 0.812 & 0.272    & 0.415 & 0.232 & \cellcolor{tabthird}4.000 & 3.846\\
        \midrule
        \multirow{4}{*}{{2D}}&LGM\cite{tang2024lgm}                 & 17.013     & 0.845      & \cellcolor{tabthird}0.033    & \cellcolor{tabsecond}61.66     & \cellcolor{tabsecond}0.45      & \cellcolor{tabthird}0.819 &0.238    & 0.431 & 0.308 & 3.538 & \cellcolor{tabthird}4.000\\
        &Zero123Plus\cite{shi2023zero123++}            & 15.787     & 0.827      & 0.037    & 82.80     & 0.82 & \cellcolor{tabsecond}0.825 & \cellcolor{tabsecond}0.165 & \cellcolor{tabsecond}0.525 & \cellcolor{tabsecond}0.409& 2.088 & 2.154\\
        &Era3D~\cite{liera3d} & 14.993  & 0.830 & 0.040 & 90.35 & 0.92 & 0.811& 0.229 & 0.436 & 0.256 & 3.846 & 2.418\\
        &SV3D\cite{voleti2024sv3d}                & \cellcolor{tabthird}17.772     & \cellcolor{tabthird}0.863      & 0.034    & \cellcolor{tabthird}74.86     & \cellcolor{tabthird}0.80    & 0.816 & 0.200    & 0.418 & \cellcolor{tabthird}0.311 & \cellcolor{tabsecond}4.231 & 2.923\\
        \midrule
        2D-3D &GSV3D                & \cellcolor{tabfirst}20.390     & \cellcolor{tabfirst}0.884     & \cellcolor{tabfirst}0.023    & \cellcolor{tabfirst}50.79     & \cellcolor{tabfirst}0.21      & \cellcolor{tabfirst}0.839 & \cellcolor{tabfirst}0.100    & \cellcolor{tabfirst}0.721 & \cellcolor{tabfirst}0.661 & \cellcolor{tabfirst}6.308 &  \cellcolor{tabfirst}6.692\\
\bottomrule
\end{tabular}}
}
\vspace{-6pt}
\caption{Quantitative comparisons on GSO dataset~\cite{downs2022google}.
GA and TGS are abbreviations for GaussianAnything and TriplaneGaussian, respectively.
GaussianAnything and TriplaneGaussians are 3D methods, while LGM, Zero123Plus, Era3D, and SV3D are 2D methods.
The second and third-best performances are highlighted in red, orange, and yellow.
Our GSV3D achieves the best performance on the GSO dataset.}
\vspace{-8pt}
\label{tab:compare}
\end{table*}

\vspace{4pt}
\noindent\textbf{Baselines.}
We compare our model against the following state-of-the-art methods: GaussianAnything~\cite{lan2024ga} (GA), TriplaneGaussian~\cite{zou2024triplane} (TGS), LGM~\cite{tang2024lgm}, Zero123Plus~\cite{shi2023zero123++}, Era3D~\cite{liera3d}, and SV3D~\cite{voleti2024sv3d}. 
Among these methods, GaussianAnything and TriplaneGaussian can be considered as 3D methods. 
They generate 3D representations without the intermediate step of estimating 2D images and are trained from scratch on 3D datasets.
In contrast, LGM, Zero123Plus, Era3D, and SV3D are regarded as 2D methods.
These methods are typically based on strong pre-trained 2D Diffusion models to help construct 3D representations.

\subsection{Evalutaion on 3D Generation}
\noindent\textbf{Appearance Quality.} From Figure~\ref{fig:compare}, it can be observed that the appearance quality of 3D methods is inferior.
Specifically, the 3D models generated by GaussianAnything exhibit color deviations from the original images (as seen in the first and second examples), while the results produced by TriplaneGaussian are blurry (as shown in the fourth example).
These issues are likely attributed to the limited dataset diversity used in training 3D methods, which hampers their ability to generate realistic appearances.

In contrast, 2D methods~\cite{tang2024lgm, shi2023zero123++, liera3d, voleti2024sv3d} demonstrate superior performance in terms of appearance generation.
The results exhibit richer and more authentic details, as shown in the third example, where the 2D methods capture more intricate folds on the horse's back.
However, due to poor image consistency, the reconstructed results of these images suffer from blurry and ghosting artifacts, which degrade the evaluation metrics, as shown in Table~\ref{tab:compare}.
In addition, Era3D performs poorly due to its design choice of generating images with an elevation of 0 to reduce computational costs.
While this approach saves attention computation, it results in inaccurate reconstructions of the unseen parts of objects and produces fragmented outputs, as demonstrated in the first and second examples.
Our method inherits the generative advantages of 2D methods while overcoming their inconsistency issues, thus achieving the best results on appearance metrics.

\vspace{4pt}
\noindent\textbf{Geometry Quality.} 
As can be seen from Figure~\ref{fig:compare}, 3D methods~\cite{lan2024ga, zou2024triplane} tend to generate intact objects by directly modeling 3D structures, avoiding the information loss inherent in 2D-to-3D indirect representations.
However, TriplaneGaussian suffers from the issue of sparse point cloud density when the object is relatively complex, which leads to holes or overly smoothness in the reconstructed mesh, as shown in the fourth and fifth examples.
On the other hand, while GaussianAnything can generate denser point clouds, its fidelity is limited, as shown in the fourth and fifth examples.
This limitation stems from the restricted diversity of the training data, which hinders the model's ability to generalize well to unseen or complex scenarios.
2D methods~\cite{tang2024lgm, shi2023zero123++, liera3d, voleti2024sv3d} suffer from geometric inconsistencies, leading to distortions and holes in the results, as shown in the second example.
Our method, constrained by explicit 3D representations, achieves better geometric quality.

\noindent\textbf{User study.} We conduct a user study to evaluate our method against existing approaches. 
For a collection of 21 images, we render 360-degree rotating videos of the 3D representations generated from these methods.
There are in total 322 videos for evaluation in our user study.
Each volunteer is presented with 10 samples from each of the following methods: GaussianAnything, TriplaneGaussian, LGM, Zero123Plus, Era3D, SV3D, and GSV3D, and asked to rank in two aspects: geometry quality and appearance quality.
The top-ranked method receives 7 points and the lowest-ranked method receives 1 point. 
We collect results from 40 volunteers and get 2800 valid scores in total.
As shown in Table~\ref{tab:compare}, our method receives the highest score from participants in terms of both appearance and geometry.

\begin{figure}[t]
  \centering
   \includegraphics[width=1\linewidth]{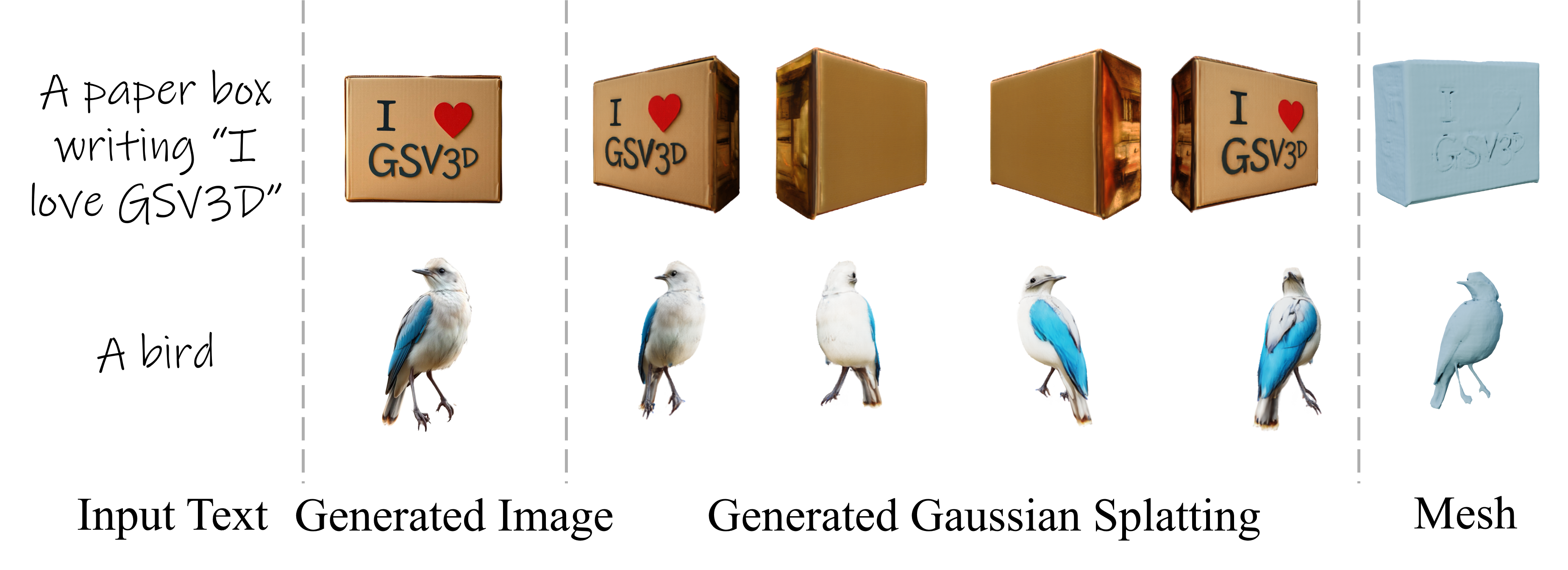}
   \vspace{-24pt}
   \caption{Examples of using GSV3D for text-to-image-to-3D generation.}
   \label{fig:text-3d}
   \vspace{-8pt}
\end{figure}

\vspace{4pt}
\noindent
\textbf{Text-to-Image-to-3D.}
By integrating advanced text-to-image models (e.g., Stable Diffusion~\cite{rombach2022high}, FLUX~\cite{flux2024}) into our GSV3D, we are capable of generating 3D models directly from textual descriptions, as shown in Figure~\ref{fig:text-3d}.

\begin{figure}[t]
  \centering
   \includegraphics[width=1\linewidth]{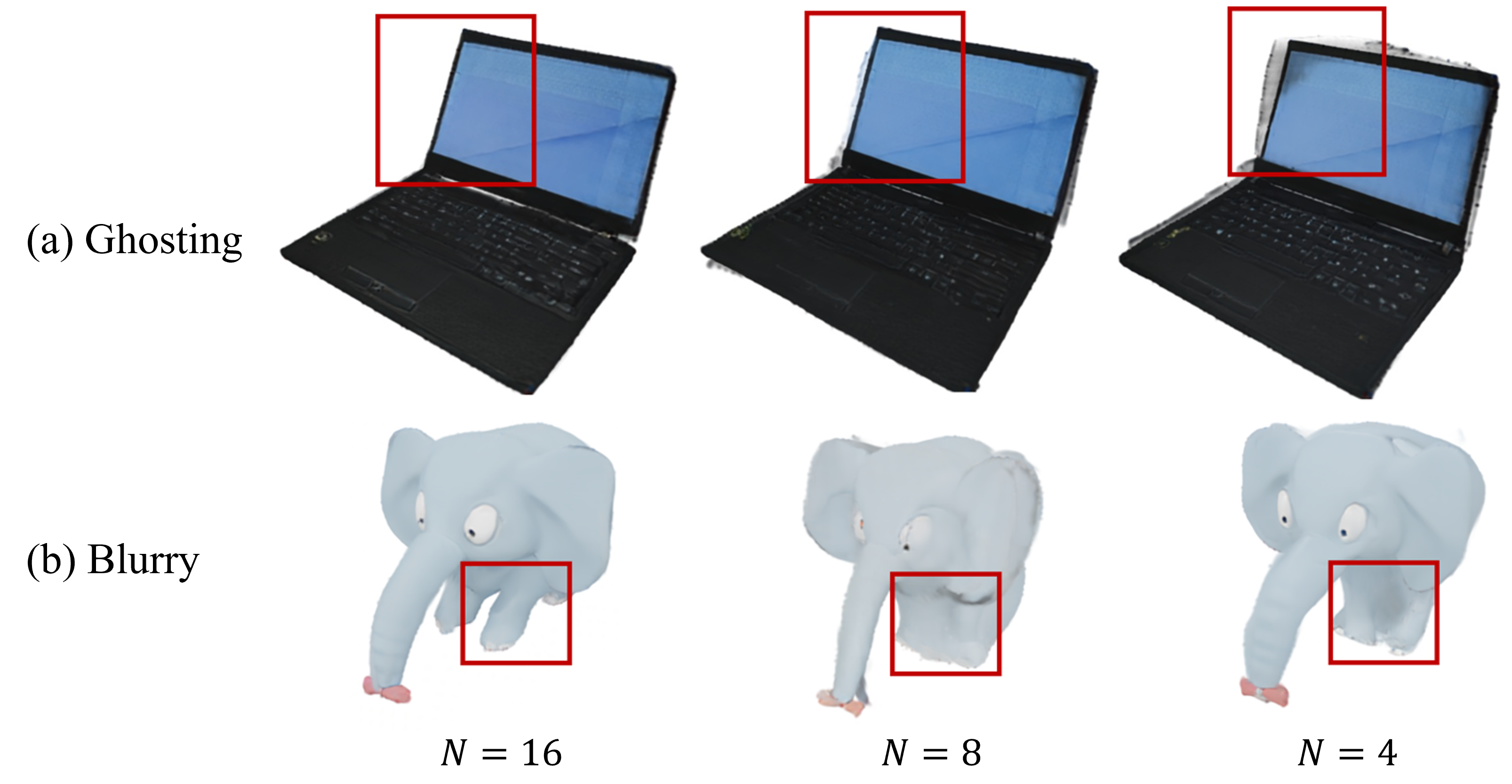}
   \vspace{-24pt}
   \caption{Visual comparison for the number of frames \(N\) in the multi-view latents generated by denoising UNet \(\epsilon_\theta\).
   Reducing the number of frames weakens generative capability and 3D constraints, leading to ghosting and blurry regions in the outputs of GSV3D.}
   \vspace{-12pt}
   \label{fig:ablation}
\end{figure}

\subsection{Ablation Studies}
In the ablation studies, all models are trained following the training procedures described previously.

\vspace{4pt}
\noindent\textbf{Effect of Geometric Distillation.} We evaluate the effectiveness of the RGB loss \(\mathcal{L_{\text{rgb}}}\) and depth loss \(\mathcal{L}_{\text{depth}}\), as mentioned in Section~\ref{sec:svd-gs}, introduced by the 3D-aware branch in the geometric distillation process.
Table~\ref{tab:ablation_studies} presents the results of ablation studies conducted by distilling our GSV3D framework with different loss functions.
The first row represents a version distilled without the RGB loss, while the second row represents a version distilled without the depth loss.
The performance drops in both cases, underscoring the importance of supervising the 3D transition from latents through both RGB and depth constraints. 

\begin{table}
    \centering
    \resizebox{\columnwidth}{!}{
    \begin{tabular}{lccccc}
        \toprule
        Setting  & PSNR↑ & SSIM↑ & LPIPS↓ & FID↓ & KID(\%)↓\\        
        \midrule
        \textit{Geometric Distillation} & & & & & \\
        \quad w/o RGB loss             & 19.551 & 0.872 & 0.027 & 53.89 & 0.27 \\
        \quad w/o Depth loss           & 19.826 & 0.877 & 0.025 & 53.20 & 0.32 \\
        \midrule
        \textit{Number of Frames} & & & & & \\
        \quad N = 8                & 19.733    & 0.877   & 0.026     & 55.49  & 0.34    \\
        \quad N = 4                & 19.548 & 0.874 & 0.027 & 57.85 & 0.42 \\
        \midrule
        \textit{DINO encoder} \\
        w/o DINO encoder & 19.508 & 0.876 & 0.027 & 55.20 & 0.30\\
        \midrule
        \textbf{Full Model (GSV3D)} & \textbf{20.390} & \textbf{0.884} & \textbf{0.023} & \textbf{50.79} & \textbf{0.21} \\
        \bottomrule
    \end{tabular}
    }
    \vspace{-6pt}
    \caption{Ablation studies on the losses used in geometric distillation, the number of frames \(N\) in the multi-view latents generated by multi-view diffusion in GSV3D, and the use of the DINO encoder.
    The full model consists of RGB loss, depth loss, a configuration of $N=16$, and a DINO encoder.
    All experiments are conducted on the GSO dataset.
    }
    \vspace{-8pt}
    \label{tab:ablation_studies}
\end{table}

\vspace{4pt}
\noindent
\textbf{Effect of the Number of Frames in the Latents Generated by GSV3D.}
To investigate the impact of the number of frames \(N\) in the multi-view latents generated by the denoising UNet \(\epsilon_\theta\) in GSV3D, as mentioned in Section~\ref{sec:svd}, we conducted experiments comparing three settings to train the GSV3D, with \(N\) set to 4, 8, and 16 frames. 
The results in Table~\ref{tab:ablation_studies} demonstrate that the setting $N=16$ outperforms the other settings in terms of overall performance. 
This improvement can be attributed to the higher number of overlapping regions between frames in the setting $N=16$, which helps maintain consistency across views and provides a clear render result without ghosting, as shown in Figure~\ref{fig:ablation}~(a).
In contrast, the setting $N=4$ may leave some areas uncovered, leading to blurry results in the final generation, as shown in Figure~\ref{fig:ablation}~(b).
Based on these results, we choose $N=16$ as the final configuration for our model.

\vspace{4pt}
\noindent
\textbf{Effect of DINO encoder in Gaussian Splatting Decoder.}
We evaluate the impact of the DINO encoder within the Gaussian Splatting Decoder, as outlined in Section~\ref{sec:gs-decoder}.
By removing the DINO encoder from the Gaussian Splatting Decoder and utilizing the modified model to distill GSV3D, we observe a performance drop, as demonstrated in Table~\ref{tab:ablation_studies}. 
This underscores the essential role of the DINO encoder in the overall framework.

\section{Conclusion}
This paper tackles the challenge of high-quality 3D object generation from a single image by integrating 2D diffusion models with explicit 3D constraints. 
While 2D diffusion models offer diversity but lack geometric consistency, and 3D diffusion models face data limitations, our hybrid approach bridges this gap.
Leveraging the Stable Video Diffusion with Gaussian Splatting, we transform multi-view latents into coherent 3D representations, enhancing consistency and fidelity. 
Extensive experiments show the superior performance of our method and generalization capabilities in generating consistent and realistic 3D objects from 2D inputs.
{
    \small
    \bibliographystyle{ieeenat_fullname}
    \bibliography{main}
}


\end{document}